\documentclass[conference,a4paper]{IEEEtran}

\usepackage{fancyhdr}
\usepackage{amsmath}
\usepackage{amssymb}
\usepackage{amscd}
\usepackage{latexsym}
\usepackage{multirow}
\usepackage{graphics}
\usepackage{color}
\usepackage{bbm}
\usepackage{amsfonts}
\usepackage{epsfig}
\usepackage{empheq}
\usepackage{mathtools}
\usepackage{cite}
\usepackage[font=footnotesize]{caption}
\usepackage[font=footnotesize]{subcaption}
\usepackage{amsthm}
\ifCLASSINFOpdf
\else
\fi
\date{}
\newcommand{\eqdef}{\stackrel{\triangle}{=}}

\DeclareMathOperator*{\argmax}{arg\,max}
\DeclareMathOperator*{\argmin}{arg\,min}

\renewcommand{\figurename}{Fig.}
\makeatletter
\long\def\@makecaption#1#2{\renewcommand{\baselinestretch}{1}\footnotesize
\vspace{10pt}\sbox{\@tempboxa}{#1. #2}
\ifdim\wd\@tempboxa > \linewidth { #1. #2\par}
\else\begin{center}#1. #2\end{center}\fi}
\makeatother

\newtheoremstyle{mynewtheorem}
{3pt}
{3pt}
{}
{}
{\itshape}
{:}
{.5em}
{\thmname{#1}\thmnumber{ #2}}%

\theoremstyle{mynewtheorem}
\newtheorem{myprop}{Proposition}

\IEEEoverridecommandlockouts

\begin{titlepage}

\title{Some Options for $L_1$-subspace Signal Processing$^*$\thanks{$^*$This work was supported in part by  the U.S. National Science Foundation under Grant CNS-1117121 and the Ministry of Education of Greece under Thales Program Grant MIS-379418-DISCO.}}

\author{%
\IEEEauthorblockN{Panos P. Markopoulos}
\IEEEauthorblockA{Electrical Engineering Dept.\\
State University of New York at Buffalo\\
Buffalo,  NY 14260\\
Email: \texttt{pmarkopo@buffalo.edu}\vspace*{-1cm}}
\and
\IEEEauthorblockN{George N. Karystinos}
\IEEEauthorblockA{Electronic and Computer  Engineering Dept.\\
Technical University of Crete\\
Chania, 73100, Greece\\
Email: \texttt{karystinos@telecom.tuc.gr}\vspace*{-1cm}}
\and
\IEEEauthorblockN{Dimitris A. Pados$^\dag$\thanks{$^\dag$Corresponding author.}}
\IEEEauthorblockA{Electrical Engineering Dept.\\
State University of New York at Buffalo\\
Buffalo,  NY 14260\\
Email: \texttt{pados@buffalo.edu}}\vspace*{-1cm}}

\end{titlepage}

\begin{document}

\maketitle
\thispagestyle{fancy}
\begin{abstract}
We describe ways to define and calculate $L_1$-norm signal subspaces which are less sensitive to outlying data than $L_2$-calculated subspaces.
We focus on the computation of the $L_1$ maximum-projection principal component of a data matrix containing $N$ signal samples of dimension $D$ and conclude that the general problem is formally NP-hard in asymptotically large $N$, $D$.
We prove, however, that the case of engineering interest of fixed dimension $D$ and asymptotically large sample support $N$ is not and we present an optimal algorithm of complexity $\mathcal O(N^D)$.
We generalize to multiple $L_1$-max-projection components and present an explicit optimal $L_1$ subspace calculation algorithm in the form of matrix nuclear-norm evaluations.
We conclude with illustrations of $L_1$-subspace signal processing in the fields of data dimensionality reduction and direction-of-arrival estimation.
\end{abstract}

\section{Introduction}

Subspace signal processing theory and practice rely, conventionally, on the familiar $L_2$-norm based singular-value decomposition (SVD) of the data matrix.
The SVD solution traces its origin to the fundamental problem of $L_2$-norm low-rank matrix approximation, which is equivalent to the problem of maximum $L_2$-norm orthonormal data projection with as many projection (``principal'') components as the desired low-rank value \cite{Golub1996}.
Practitioners have long observed, however, that $L_2$-norm principal component analysis (PCA) is sensitive to the presence of outlier values in the data matrix, that is, values that are away from the nominal distribution data, appear only few times in the data matrix, and are not to appear again under normal system operation upon design.
This paper makes a case for $L_1$-subspace signal processing.
Interestingly, in contrast to $L_2$, subspace decomposition under the $L_1$ error minimization criterion and the $L_1$ projection maximization criterion are not the same.
A line of recent research pursues calculation of $L_1$ principal components under error minimization [\citen{Ke2005}] or projection maximization [\citen{Kwak2008}], [\citen{Nie2011}].%
\footnote{A combined $L_1$/$L_2$-norm approach has been followed in [\citen{Ding2006}].}
No algorithm has appeared so far with guaranteed convergence to the criterion-optimal subspace and no upper bounds are known on the expended computational effort.

In this present work, given any data matrix ${\mathbf X}\in \mathbb R^{D \times N}$ of $N$ signal samples of dimension $D$, we show that the general problem of finding the maximum $L_1$-projection principal component of $\mathbf X$ is formally NP-hard for asymptotically large $N$, $D$.
We prove, however, that the case of engineering interest of fixed given dimension $D$ is not NP-hard.
In particular, for the case where $N<D$, we present in explicit form an algorithm to find the optimal component with computational cost $2^N$.
For the case where the sample support exceeds the data dimension ($N \geq D$) --which is arguably of more interest in signal processing applications-- we present an algorithm that computes the $L_1$-optimal principal component with complexity $\mathcal O\big(N^{\text{rank}(\mathbf X)}\big)$, $\text{rank}(\mathbf X) \leq D$.
We generalize the effort to the problem of calculating multiple $L_1$ components (necessarily a joint computational problem) and present an explicit optimal algorithm for multi-component subspace design in the form of matrix nuclear-norm maximization.

\section{Problem Statement}

Consider $N$ real-valued measurements $\mathbf x_1, \mathbf x_2, \cdots, \mathbf x_N$ of dimension $D$ that form the $D\times N$ data matrix
\begin{equation}
{\bf X}=[{\bf x}_1\;\;{\bf x}_2\;\ldots\;{\bf x}_N].
\end{equation}
We are interested in describing (approximating) the data matrix ${\bf X}$ by a rank-$K$ product ${\bf R}{\bf S}^T$ where ${\bf R} \in \mathbb R^{D \times K}$, ${\bf S} \in \mathbb R^{N \times K}$, $K \leq D$, in the form of Problem $\mathcal P_1^{L_2}$  defined below,
\begin{equation}
{\mathcal P}^{L_2}_1:\;\;\;\;\left({\bf R}_{L_2},{\bf S}_{L_2}\right)=\argmin_{{\bf R}\in{\mathbb R}^{D\times K},\;{\bf S}\in{\mathbb R}^{N\times K}}\left\|{\bf X}-{\bf R}{\bf S}^T\right\|_2
\label{eq:RS}
\end{equation}
where $\|{\bf A}\|_2=\sqrt{\sum_{i,j}|A_{i,j}|^2}$ is the $L_2$ matrix norm (Frobenius) of  matrix $\mathbf A$ with elements $A_{i,j}$.
By the Projection Theorem \cite{Golub1996}, ${\bf S}={\bf X}^T{\bf R}$ for any fixed ${\bf R}$, $\mathbf R^T \mathbf R = \mathbf I_K$.
Hence, we obtain the equivalent problem
\begin{equation}
{\mathcal P}^{L_2}_2:\;\;\;\;{\bf R}_{L_2}=\argmin_{
{\bf R}\in{\mathbb R}^{D\times K},\,{\bf R}^T{\bf R}={\bf I}_K}\left\|{\bf X}-{\bf R}{\bf R}^T{\bf X}\right\|_2
\label{eq:RR}
\end{equation}
frequently referred to as left-side $K$-SVD.
Since $\left\|{\bf A}\right\|_2^2=\text{tr}\left({\bf A}^T{\bf A}\right)$ where $\text{tr}(\cdot)$ denotes the trace of a matrix,  ${\mathcal P}^{L_2}_2$ is also equivalent to   $L_2$ projection (energy) maximization,
\begin{equation}
{\mathcal P}^{L_2}_3:\;\;\;\;{\bf R}_{L_2}=\argmax_{{\bf R}\in{\mathbb R}^{D\times K},\,{\bf R}^T{\bf R}={\bf I}_K}\left\|{\bf X}^T{\bf R}\right\|_2.
\label{eq:R}
\end{equation}
Note that, if $K<D$ and we possess the solution ${\bf R}_{L_2}^{(K)}$ for $K$ singular/eigen-vectors  in (\ref{eq:RS}), (\ref{eq:RR}), (\ref{eq:R}), then the solution for rank $K+1$ is derived readily by ${\bf R}_{L_2}^{(K+1)}= [{\bf R}_{L_2}^{(K)}\;\;{\bf r}_{L_2}^{(K+1)} ]$ with
$
{\bf r}_{L_2}^{(K+1)}=\argmax_{{\bf r}\in{\mathbb R}^D,\;\left\|{\bf r}\right\|_2=1} \|{\bf X}^T  ({\bf I}_D-{\bf R}_{L_2}^{(K)}{{\bf R}_{L_2}^{(K)}}^T ){\bf r} \|_2.
$
This is known as the PCA scalability property.

By minimizing the sum of squared errors, $L_2$ principal component calculation becomes sensitive to extreme error value occurrences caused by  the presence of outlier measurements in the data matrix.
Motivated by this observed drawback of $L_2$ subspace signal processing, in this work we study and  pursue subspace-decomposition approaches that are based on the $L_1$ norm,
$\left\|{\bf A}\right\|_1=\sum_{i,j}\left|A_{i,j}\right|$.
We  may ``translate" the three equivalent $L_2$ optimization problems (\ref{eq:RS}), (\ref{eq:RR}), (\ref{eq:R}) to new problems that utilize the $L_1$ norm as follows,
\begin{align}
&{\mathcal P}^{L_1}_1:\; \left({\bf R}_{L_1},{\bf S}_{L_1}\right)= \hspace{-0.3cm} \argmin_{ \begin{smallmatrix} {\bf R}\in{\mathbb R}^{D\times K},\;{\bf R}^T{\bf R}={\bf I}_K, \\ {\bf S}\in{\mathbb R}^{N\times K} \end{smallmatrix} } \hspace{-0.3cm}  \left\|{\bf X}-{\bf R}{\bf S}^T\right\|_1,\\
\label{eq:RSL1}
&{\mathcal P}^{L_1}_2:\; {\bf R}_{L_1}=\argmin_{{\bf R}\in{\mathbb R}^{D\times K},\;  {\bf R}^T{\bf R}={\bf I}_K}\left\|{\bf X}-{\bf R}{\bf R}^T{\bf X}\right\|_1,\\
&{\mathcal P}^{L_1}_3:\; {\bf R}_{L_1}=\argmax_{{\bf R}\in{\mathbb R}^{D\times K}, \;  {\bf R}^T{\bf R}={\bf I}_K}\left\|{\bf X}^T{\bf R}\right\|_1.
\label{eq:RL1}
\end{align}
A few comments appear useful at this point: 
(i) Under the $L_1$ norm, the three optimization problems ${\mathcal P}^{L_1}_1$, ${\mathcal P}^{L_1}_2$, and ${\mathcal P}^{L_1}_3$ are no longer equivalent.
(ii) Under $L_1$, the PCA scalability property does not hold (due to loss of the Projection Theorem).
(iii) Even for reduction to a single dimension (rank $K=1$ approximation), the three problems are difficult to solve.

In this present work, we focus exclusively on ${\mathcal P}^{L_1}_3$.

\section{The $L_1$-norm Principal Component}
\label{sec:OneComponent}

In this section, we concentrate on the calculation of the $L_1$-maximum-projection component of a data matrix $\mathbf X \in \mathbb R^{D \times N}$ (Problem $\mathcal P_3^{L_1}$ in (\ref{eq:RL1}), $K=1$).
First, we  show that the problem is in general NP-hard and review briefly suboptimal techniques from the literature.
Then, we  prove that, if the data dimension $D$ is fixed, the principal $L_1$-norm component of $\mathbf X$ is in fact computable in polynomial time and present a calculation algorithm  with complexity ${\mathcal O}\left(N^{\text{rank}({\bf X})}\right)$, $\text{rank}({\bf X})\leq D$.

\subsection{The Hardness of the Problem and an Exhaustive-search Algorithm Over the Binary Field}

In Proposition \ref{prop:quad} below, we  present a fundamental property of Problem  ${\mathcal P}^{L_1}_3$, $K=1$, that will lead us to an efficient solution.
The proof is omitted due to lack of space and can be found in \cite{L1_TSP}.
\begin{myprop}
For any data matrix $\mathbf X \in \mathbb R^{D \times N}$, the solution to 
${\mathcal P}^{L_1}_3:{\bf r}_{L_1}=\argmax_{{\bf r}\in{\mathbb R}^D,\left\|{\bf r}\right\|_2=1}\left\|{\bf X}^T{\bf r}\right\|_1$
is given by
\begin{equation}
{\bf r}_{L_1}=\frac{{\bf X}{\bf b}_\text{opt}}{\left\|{\bf X}{\bf b}_\text{opt}\right\|_2}
\label{eq:rL1}
\end{equation}
where\begin{equation}
{\bf b}_\text{opt}=\argmax_{{\bf b}\in\{\pm1\}^N}\left\|{\bf X}{\bf b}\right\|_2=\argmax_{{\bf b}\in\{\pm1\}^N}{\bf b}^T{\bf X}^T{\bf X}{\bf b}.
\label{eq:bopt}
\end{equation}
In addition,
$\left\|{\bf X}^T {\bf r}_{L_1}\right\|_1=\left\|{\bf X}{\bf b}_\text{{opt}}\right\|_2$.
\hfill
$\Box$
\label{prop:quad}
\end{myprop}

The straightforward approach to solve (\ref{eq:bopt}) is an exhaustive search among all $2^N$ binary vectors of length $N$.
Proposition \ref{prop:NPhard} below declares that, indeed, in its general form $\mathcal P_3^{L_1}$, $K=1$, is  NP-hard for jointly asymptotically large $N, D$.
The proof can be found in \cite{L1_TSP}.
\begin{myprop}
Computation of the $L_1$ principal component of ${\bf X}\in{\mathbb R}^{D\times N}$ by maximum $L_1$-norm projection (Problem $\mathcal P_{3}^{L_1}$, $K=1$) is NP-hard in jointly asymptotic $N,D$.
\hfill
$\Box$
\label{prop:NPhard}
\end{myprop}

\subsection{Existing Approaches in Literature}

There has been a growing documented effort to calculate subspace components by $L_1$ projection maximization~[\citen{Kwak2008}], [\citen{Nie2011}].
For $K=1$, both algorithms in~\cite{Kwak2008}, \cite{Nie2011} are identical and can  be described by the simple single iteration
\begin{equation}
{\bf b}^{(i+1)}=\text{sgn}\left({\bf X}^T{\bf X}{\bf b}^{(i)}\right),\;\;\;i=1,2,\ldots,
\label{eq:Kwak3}
\end{equation}
for the computation of ${\bf b}_\text{opt}$ in~(\ref{eq:bopt}).
Equation \eqref{eq:Kwak3}, however, does not guarantee convergence to the $L_1$-optimal component solution (convergence to one of the many local maxima may be observed).
In the following section, we present for the first time in the literature an optimal algorithm to calculate the $ L_1$ principal component of a data matrix with complexity polynomial in the sample support $N$ when the data dimension $D$ is fixed.

\subsection{Computation of the $L_1$ Principal Component in Polynomial Time}

In the following, we show   that, if $D$ is fixed, then  computation of ${\bf r}_{L_1}$ is no longer NP-hard (in $N$).
We state our result in the form of Proposition \ref{prop:polynomial} below.
\begin{myprop}
For any fixed data dimension $D$,  computation of the $L_1$ principal component of ${\bf X}\in{\mathbb R}^{D\times N}$ has complexity ${\mathcal O}\left(N^{\text{{rank}}({\bf X})}\right)$, $\text{{rank}}({\bf X})\leq D$.
\hfill
$\Box$
\label{prop:polynomial}
\end{myprop}
By Proposition \ref{prop:NPhard}, computation of the $ L_1$ principal component of $\mathbf X$ is equivalent to computation of $\mathbf b_{\text{opt}}$ in (\ref{eq:bopt}).
To prove Proposition \ref{prop:polynomial}, we will then prove that ${\bf b}_\text{opt}$ can be computed with complexity ${\mathcal O}\left(N^{\text{rank}({\bf X})}\right)$.
We begin our developments by defining
\begin{align}
d\eqdef\text{rank}({\bf X})\leq D.
\end{align}
Then, ${\bf X}^T{\bf X}$ has also rank $d$ and can be  decomposed by
\begin{align}
{\bf X}^T{\bf X}={\bf Q}{\bf Q}^T,\; {\bf Q}_{N\times d}=\left[{\bf q}_1\;{\bf q}_2\;\ldots\;{\bf q}_d\right],\;{\bf q}_i^T{\bf q}_j=0,\;i\neq j,
\label{eq:QQ}
\end{align}
where ${\bf q}_1$, ${\bf q}_2$, $\ldots$ , ${\bf q}_d$ are the $d$ eigenvalue-weighted eigenvectors of ${\bf X}^T{\bf X}$ with nonzero eigenvalue.
By~(\ref{eq:bopt}),
\begin{equation}
{\bf b}_\text{opt}=\argmax_{{\bf b}\in\{\pm1\}^N}{\bf b}^T{\bf Q}{\bf Q}^T{\bf b}=\argmax_{{\bf b}\in\{\pm1\}^N}\left\|{\bf Q}^T{\bf b}\right\|_2.
\label{eq:Qb}
\end{equation}

For the case  $N<D$, the optimal binary vector ${\bf b}_\text{opt}$ can be obtained directly from~(\ref{eq:Qb}) by an exhaustive search among all $2^N$ binary vectors ${\bf b}\in\{\pm1\}^N$.
Therefore, we can design the $L_1$-optimal principal component ${\bf r}_{L_1}$ with computational cost $2^N<2^D={\mathcal O}(1)$.
For the case where the sample support exceeds the data dimension ($N\geq D$) -which is arguably of higher interest in signal processing applications- we find it useful in terms of both theory and practice to present  our developments separately for data rank $d=1$, $d=2$, and $2 < d \leq D$.\\
\emph{1) Case $d=1$:}
If the data matrix has rank  $d=1$, then ${\bf Q}={\bf q}_1$ and~(\ref{eq:Qb}) becomes
\begin{equation}
{\bf b}_\text{opt}=\argmax_{{\bf b}\in\{\pm1\}^N}\left|{\bf q}_1^T{\bf b}\right|=\text{sgn}\left({\bf q}_1\right).
\end{equation}
By~(\ref{eq:rL1}), the $L_1$-optimal principal component is
\begin{equation}
{\bf r}_{L_1}=\frac{{\bf X}\,\text{sgn}\left({\bf q}_1\right)}{\left\|{\bf X}\,\text{sgn}\left({\bf q}_1\right)\right\|_2}
\label{dequals1}
\end{equation}
designed with complexity ${\mathcal O}\left(N\right)$.
It is of notable practical importance to observe  at this point that even when $\mathbf X$ is not of true rank one, \eqref{dequals1} presents us with a quality, trivially calculated approximation of the $L_1$ principal component of $\mathbf X$:
Calculate the $L_2$ principal component $\mathbf q_1$ of the $N \times N$ matrix $\mathbf X^T \mathbf X$, quantize to $\text{sgn}(\mathbf q_1)$, and project and normalize to obtain $\mathbf r_{L_1} \simeq \mathbf X\,\text{sgn}(\mathbf q_1) / \| \mathbf X\,\text{sgn}(\mathbf q_1)  \|_2$.\\
\emph{2) Case $d=2$:}
If $d=2$, then ${\bf Q}=\left[{\bf q}_1\;\;{\bf q}_2\right]$ and~(\ref{eq:Qb}) becomes
\begin{equation}
{\bf b}_\text{opt}=\argmax_{{\bf b}\in\{\pm1\}^N}\left\{\left({\bf q}_1^T{\bf b}\right)^2+\left({\bf q}_2^T{\bf b}\right)^2\right\}.
\label{eq:rank2}
\end{equation}
The binary optimization problem \eqref{eq:rank2} was seen and solved for the first time in \cite{KP} by the auxiliary-angle method \cite{mackenthun} with complexity ${\mathcal O}\left(N\log_2N\right)$.
Due to lack of space, we omit the specifics of the Case $d=2$ and move directly to the general case $2 \leq  d \leq D$.

\emph{3) Case $2 \leq d \leq D$:}
If $d \geq 2$, we design the $L_1$-optimal principal component of ${\bf X}$ with complexity ${\mathcal O}\left(N^d\right)$ by considering the multiple-auxiliary-angle approach that was presented in~\cite{KL} as a generalization of the work in~\cite{KP}.

Consider a unit vector ${\bf c}\in{\mathbbm R}^d$. By  Cauchy-Schwartz,  for any ${\bf a}\in{\mathbbm R}^d$,
\begin{align}
{\bf a}^T{\bf c}\leq\left\|{\bf a}\right\|_2\left\|{\bf c}\right\|_2=\left\|{\bf a}\right\|_2
\end{align}
with equality if and only if ${\bf c}$ is codirectional with ${\bf a}$.
Then,
\begin{align}
\max_{{\bf c}\in{\mathbbm R}^d,\;\left\|{\bf c}\right\|=1}{\bf a}^T{\bf c}=\left\|{\bf a}\right\|_2.
\label{eq:ca}
\end{align}
By~(\ref{eq:ca}), the optimization problem in~(\ref{eq:Qb}) becomes
\begin{align}
 \max_{{\bf b}\in\{\pm1\}^N}\left\|{\bf Q}^T{\bf b}\right\|_2 &=\max_{{\bf b}\in\{\pm1\}^N}\max_{{\bf c}\in{\mathbbm R}^d,\; \left\|{\bf c}\right\|=1}{\bf b}^T{\bf Q}{\bf c} \nonumber \\
&=\max_{{\bf c}\in{\mathbbm R}^d,\; \left\|{\bf c}\right\|=1}\max_{{\bf b}\in\{\pm1\}^N}{\bf b}^T{\bf Q}{\bf c}.
 \label{eq:maxmax}
\end{align}
For every ${\bf c}\in{\mathbbm R}^d$,  inner maximization in~(\ref{eq:maxmax}) is solved by the binary vector
\begin{align}
{\bf b}({\bf c}) = \text{sgn}({\bf Q}{\bf c}),
\label{eq:bc}
\end{align}
which is obtained with complexity ${\mathcal O}(N)$.
Then, by~(\ref{eq:maxmax}), the solution to the original problem in~(\ref{eq:Qb}) is met if we collect all  binary vectors ${\bf b}({\bf c})$ returned as ${\bf c}$ scans the unit-radius $d$-dimensional hypersphere.
That is, ${\bf b}_\text{opt}$ in~(\ref{eq:Qb}) is in%
\footnote{The $d$th element of vector $\mathbf c$, $c_d$, can be set nonnegative without loss of optimality, because, for any given $\mathbf c$, $\| \mathbf c\|_2=1$, the binary vectors $\mathbf b(\mathbf c)$ and $\mathbf b(\text{sgn}(c_d)\mathbf c)$ result to the same metric value in \eqref{eq:Qb}.}
\begin{align}
{\mathcal S}\eqdef\hspace{-.5cm}\bigcup_{{\bf c}\in{\mathbbm R}^d,\,\left\|{\bf c}\right\|_2=1, \,c_d\geq0}\hspace{-.5cm}{\bf b}({\bf c}).
\label{eq:S1}
\end{align}

Two fundamental questions for the computational problem under consideration are what the size (cardinality) of set ${\mathcal S}$ is and  how much computational effort is expended to form ${\mathcal S}$.

The candidate vector set ${\mathcal S}$ has cardinality $|{\mathcal S}|=\sum_{g=0}^{d-1}\binom{N-1}{g}={\mathcal O}\left(N^{d-1}\right)$ and it suffices to solve
\begin{align}
{\bf Q}_{\mathcal I,:}{\bf c} ={\bf 0} 
\label{eq:Qc}
\end{align}
for every $\mathcal I \subset \{1, 2, \cdots, N\}$, $|\mathcal I |=d-1$ (i.e., $\mathbf Q_{\mathcal I, :}$ contains any $d-1$ rows of $\mathbf Q$).
The solution to~(\ref{eq:Qc}) is the unit vector in the null space of the $(d-1)\times d$ matrix ${\bf Q}_{\mathcal I,:}$.%
\footnote{If ${\bf Q}_{\mathcal I,:}$ is full-rank, then its null space has rank $1$ and ${\bf c} $ is uniquely determined (within a sign ambiguity which is resolved by $c_d\geq0$).
If, instead, ${\bf Q}_{\mathcal I,:}$ is rank-deficient, then the intersection of the $d-1$ hypersurfaces (i.e., the solution of~(\ref{eq:Qc})) is a $p$-manifold (with $p\geq1$) in the $(d-1)$-dimensional space and does not generate  new binary vectors of interest.
Hence, linearly dependent combinations of $d-1$ rows of ${\bf Q}$ are ignored.}
Then, the binary vectors ${\bf b}$ of interest are obtained  by
\begin{align}
\text{sgn}({\bf Q}\,{\bf c})
\label{eq:sgnQc}
\end{align}
with complexity ${\mathcal O}(N)$.
Note that~(\ref{eq:sgnQc}) presents ambiguity regarding the sign of the intersecting $d-1$ hypersurfaces (zero values).
A straightforward way to resolve the ambiguity\footnote{The algorithm of Fig.~\ref{fig:algo} uses an alternative way of resolving the sign ambiguities at the intersections of hypersurfaces which was developed in~\cite{KL} and led to the direct construction of a set ${\mathcal S}$ of size $\sum_{g=0}^{d-1}\binom{N-1}{g}={\mathcal O}(N^{d-1})$ with complexity ${\mathcal O}(N^d)$.} is to consider all $2^{d-1}$ sign combinations for the $d-1$ zero value positions.  
Since complexity ${\mathcal O}(N)$ is required to solve~(\ref{eq:sgnQc}) for each subset of $d-1$ rows of ${\bf Q}$, the overall complexity of the construction of ${\mathcal S}$ is ${\mathcal O}(N^d)$ for any given matrix ${\bf Q}_{N\times d}$.
Our complete, new algorithm for the computation of the $L_1$-optimal principal component of a rank-$d$ matrix ${\bf X}\in{\mathbb R}^{D\times N}$ that has complexity ${\mathcal O}\left(N^d\right)$ is presented in detail in Fig.~\ref{fig:algo}.%

\begin{figure} 
%
{\small
\hrule
\vspace{0.5mm}
\hrule
\vspace{1.5mm}
\noindent {\bf The Optimal $L_1$-Principal-Component Algorithm}
\vspace{0.8mm}
\hrule
\noindent
\vspace{1.2mm}\\
\noindent \begin{tabular}{l}
\noindent {\bf Input:} $\mathbf X_{D \times N}$ data matrix\\
$\left( \mathbf U_{N \times d}, \mathbf \Sigma_{d \times d}, \mathbf V_{d\times d}\right) \leftarrow \mathrm{svd} (\mathbf X^T)$\\
$\mathbf Q_{N \times d} \leftarrow \mathbf {U \Sigma}$\\
$\mathbf B \leftarrow \mathrm{compute\_candidates} (\mathbf Q)$\\
$m_{\text{opt}} \leftarrow  {\arg \max}_m~ \mathbf B_{:,m}^T \mathbf X^T \mathbf X \mathbf B_{:,m}$\\
$\mathbf b_{\text{opt}} \leftarrow \mathbf B_{:, m_{\text{opt}}}$\\
\noindent {\bf Output:} $\mathbf r_{L_1} \leftarrow {\mathbf X \mathbf b_{\text{opt}}}/{\|\mathbf X \mathbf b_{\text{opt}}\|_2}$\\
\end{tabular}
\vspace{1mm}\\
\hrule
\vspace{1.2mm}
\noindent {Function \emph{compute\_candidates}}
\vspace{0.8mm}
\hrule
\noindent
\vspace{1.2mm}\\
\noindent \begin{tabular}{l}
\noindent {\bf Input:} $\mathbf Q_{N \times m}$\\
if $m > 2$,~  $i \leftarrow 0$\\
$~~~$\noindent \begin{tabular}{l}
for $\mathcal I \subset \{1, 2, \cdots, N \}$ s.t. $| \mathcal I|=m-1$, $i \leftarrow i+1$,\\
$~~~$\noindent \begin{tabular}{l}
$\bar{\mathbf Q}_{(m-1) \times m} \leftarrow \mathbf Q_{\mathcal I, :}$\\
$\mathbf c_{m \times 1} \leftarrow \mathrm{null}(\bar{\mathbf Q})$, $\mathbf c \leftarrow \mathrm{sgn}(c_m) \mathbf c$\\
$\mathbf B_{:,i} \leftarrow \mathrm{sgn}(\mathbf Q \mathbf c)$\\
for $j =1 : m-1$,\\
$~~~$\noindent \begin{tabular}{l}
$ \mathbf c_{(m-1) \times 1} \leftarrow   \mathrm{null}(\bar{\mathbf Q}_{:/j,1:m-1}) $, $\mathbf c \leftarrow  \mathrm{sgn}(c_{m-1}) \mathbf c$\\
$\mathbf B_{\mathcal I(j), i} \leftarrow \mathrm{sgn}(\bar{\mathbf Q}_{j,1:m-1} \mathbf c)$
\end{tabular}\\
\end{tabular}\\
$\mathbf B \leftarrow [\mathbf B, \mathrm{compute\_candidates}(\mathbf Q_{:,1:m-2})]$
\end{tabular}\\
elseif $m=2$,\\
$~~~~$\noindent \begin{tabular}{l}
for $i =1 : N$,\\
$~~~$\noindent \begin{tabular}{l}
$\mathbf c_{2 \times 1} \leftarrow    \mathrm{null}( \mathbf Q_{i, :})$, $\mathbf c \leftarrow  \mathrm{sgn}(c_2) \mathbf c$\\
$\mathbf B_{:,i} \leftarrow \mathrm{sgn}(\mathbf Q \mathbf c)$, $\mathbf B_{i,i} \leftarrow \mathrm{sgn}({\mathbf Q}_{i,1})$
\end{tabular}\\
\end{tabular}\\
else, $\mathbf B \leftarrow \mathrm{sgn}(\mathbf Q)$\\
\noindent {\bf Output:} $\mathbf B$\\
\end{tabular}
\vspace{0.5mm}
\hrule
\vspace{0.5mm}
\hrule
}
\caption{The optimal ${\mathcal O}(N^d)$ algorithm for the computation of the maximum $L_1$-projection component of a rank-$d$ data matrix $\mathbf X_{D \times N}$ of $N$ samples of dimension $D$.}
\vspace{-0.3cm}
\label{fig:algo}
\end{figure}

\section{Multiple $L_1$-norm Principal Components}

In this section, we switch our interest to the joint design of    $K>1$ principal $L_1$ components of a $D\times N$ matrix ${\bf X}$.

\subsection{Existing Approaches in Literature}

For the case  $K>1$,~\cite{Kwak2008} proposed to design the first $L_1$ principal component ${\bf r}_{L_1}$ by the coupled  iteration~(\ref{eq:Kwak3}) (which does not guarantee optimality) and then project the data onto the subspace that is orthogonal to ${\bf r}_{L_1}$, design the $L_1$ principal component of the projected data by the same coupled iteration, and continue similarly.
To avoid the above suboptimal greedy approach,~\cite{Nie2011} presented an iterative algorithm for the computation of ${\bf R}_{L_1}$ altogether (that is the joint computation of the $K$ principal $L_1$ components), which  does not guarantee convergence to the $L_1$-optimal subspace.

\subsection{Exact Computation of Multiple $L_1$ Principal Components}

For any $D\times K$ matrix ${\bf A}$,
\begin{align}
\max_{{\bf R}\in{\mathbbm R}^{D\times K},\,{\bf R}^T{\bf R}={\bf I}_K}\text{tr}\left({\bf R}^T{\bf A}\right)=\left\|{\bf A}\right\|_*
\label{eq:RAA}
\end{align}
where $\left\|{\bf A}\right\|_*$ denotes the nuclear norm (i.e., the sum of the singular values) of ${\bf A}$.
Maximization in~(\ref{eq:RAA}) is achieved by ${\bf R}={\bf U}{\bf V}^T$ where ${\bf U}{\bf\Sigma}{\bf V}^T$ is the ``compact'' SVD of ${\bf A}$, ${\bf U}$ and ${\bf V}$ are $D\times d$ and $K\times d$, respectively, matrices with ${\bf U}^T{\bf U}={\bf V}^T{\bf V}={\bf I}_d$, ${\bf\Sigma}$ is a nonsingular diagonal $d\times d$ matrix, and $d$ is the rank of ${\bf A}$.
This is due to the trace version of the Cauchy-Schwarz inequality~\cite{Cauchy}, according to which
\begin{align}
\text{tr}\left({\bf R}^T{\bf A}\right) 
& \leq\left\|{\bf U}{\bf\Sigma}^\frac{1}{2}\right\|_2\left\|{\bf\Sigma}^\frac{1}{2}{\bf V}^T{\bf R}^T\right\|_2=\left\|{\bf\Sigma}^\frac{1}{2}\right\|_2^2 \nonumber \\
&=\text{tr}\left({\bf\Sigma}\right)=\left\|{\bf A}\right\|_*
\label{eq:SA}
\end{align}
with equality if $\left({\bf U}{\bf\Sigma}^\frac{1}{2}\right)^T={\bf\Sigma}^\frac{1}{2}{\bf V}^T{\bf R}^T$ which is satisfied by ${\bf R}={\bf U}{\bf V}^T$.

To identify the optimal $L_1$ subspace for any number of  components $K$, we begin by presenting a property of ${\mathcal P}^{L_1}_3$ in the form of Proposition~\ref{prop:nuclear} below. 
The proof is omitted and can be found in \cite{L1_TSP}.
\begin{myprop}
For any data matrix $\mathbf X \in \mathbb R^{D \times N}$, the solution to 
${\mathcal P}^{L_1}_3: {\bf R}_{L_1}=\argmax_{{\bf R}\in{\mathbb R}^{D\times K}, \; {\bf R}^T{\bf R}={\bf I}_K}\left\|{\bf R}^T{\bf X}\right\|_1$
is given by
\begin{align}
{\bf R}_{L_1}={\bf U}{\bf V}^T
\end{align}
where ${\bf U}$ and ${\bf V}$ are the $D\times K$ and $N\times K$ matrices that consist of the $K$ highest-singular-value left and right, respectively, singular vectors of ${\bf X}{\bf B}_\text{{opt}}$ with
\begin{align}
{\bf B}_\text{{opt}}=\argmax_{{\bf B}\in\{\pm1\}^{N\times K}}\left\|{\bf X}{\bf B}\right\|_*.
\label{eq:Bopt}
\end{align}
In addition,
$\left\|{\bf R}_{L_1}^T{\bf X}\right\|_1=\left\|{\bf X}{\bf B}_\text{{opt}}\right\|_*$.
\hfill
$\Box$
\label{prop:nuclear}
\end{myprop}

By Proposition~\ref{prop:nuclear}, to find exactly the optimal $L_1$-norm projection operator ${\bf R}_{L_1}$ we can perform the following steps:
\begin{enumerate}
\item Solve~(\ref{eq:Bopt}) to obtain ${\bf B}_\text{opt}$. 
\item Perform SVD on ${\bf X}{\bf B}_\text{opt}={\bf U}{\bf\Sigma}{\bf V}^T$.
\item Return ${\bf R}_{L_1}={\bf U}_{:,1:K}{\bf V}^T$.
\end{enumerate}
Step $1$ can be executed by an exhaustive search among all $2^{NK}$ binary matrices of size $N\times K$ followed by evaluation in  the metric of interest in~(\ref{eq:Bopt}).
That is, with computational cost $\mathcal O(2^{NK})$ we identify the $L_1$-optimal $K$ principal components of ${\bf X}$.
An optimal algorithm for the computation of the $L_1$-optimal $K$ principal components of ${\bf X}$ with complexity ${\mathcal O}(N^{K\text{rank}({\bf X})-K+1})$, $\text{rank}({\bf X})\leq D$, is presented in~\cite{L1_TSP}.

\section{Experimental Studies}

%
%

\begin{figure*}
\centering
\begin{subfigure}[]{0.329\linewidth}
\centering
\includegraphics[width=\linewidth]{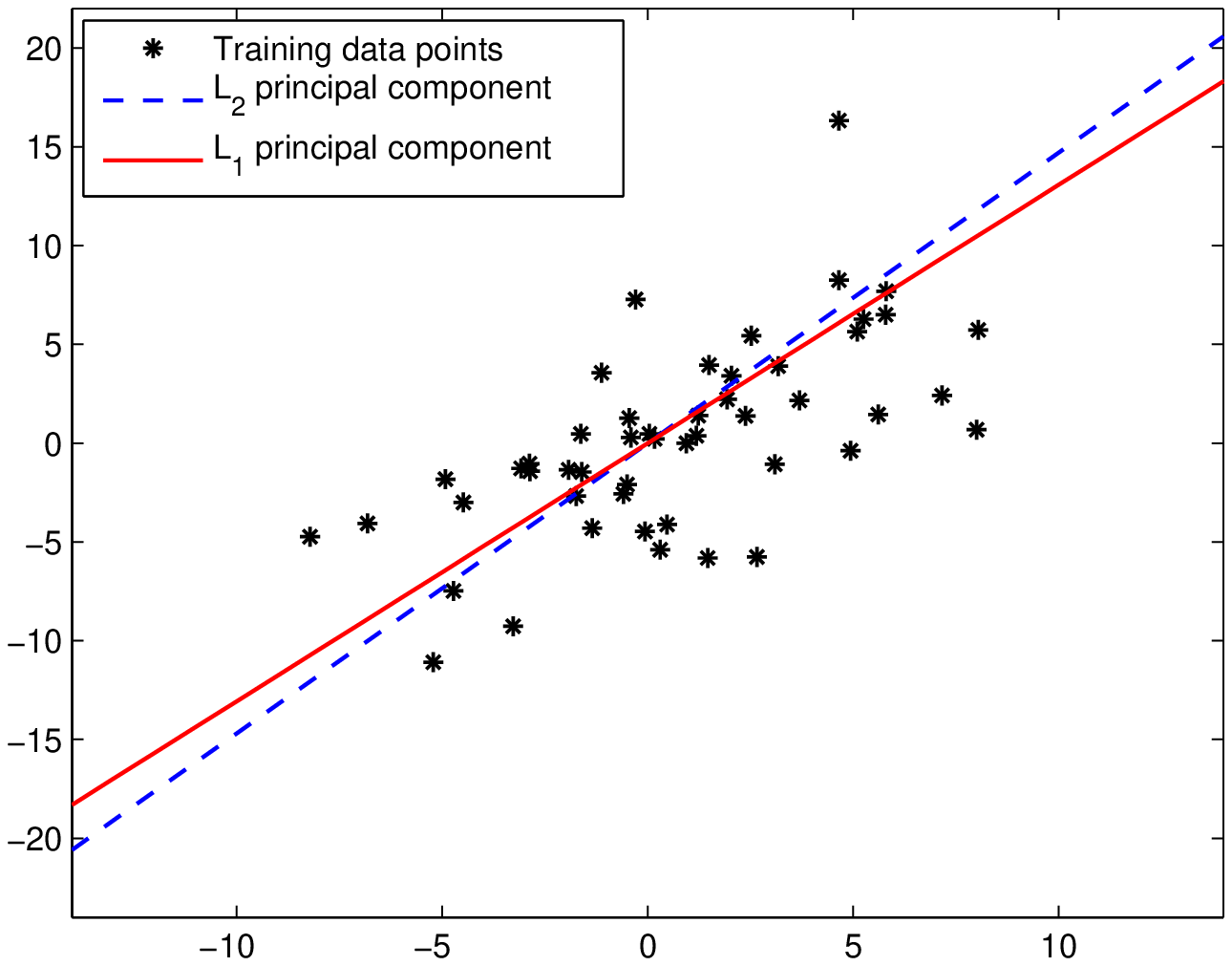}
\vspace{-0.8cm}
\caption{}
\label{fig:dr1}
\end{subfigure}
\hfill
\begin{subfigure}[]{0.329\linewidth}
\centering
\includegraphics[width=\linewidth]{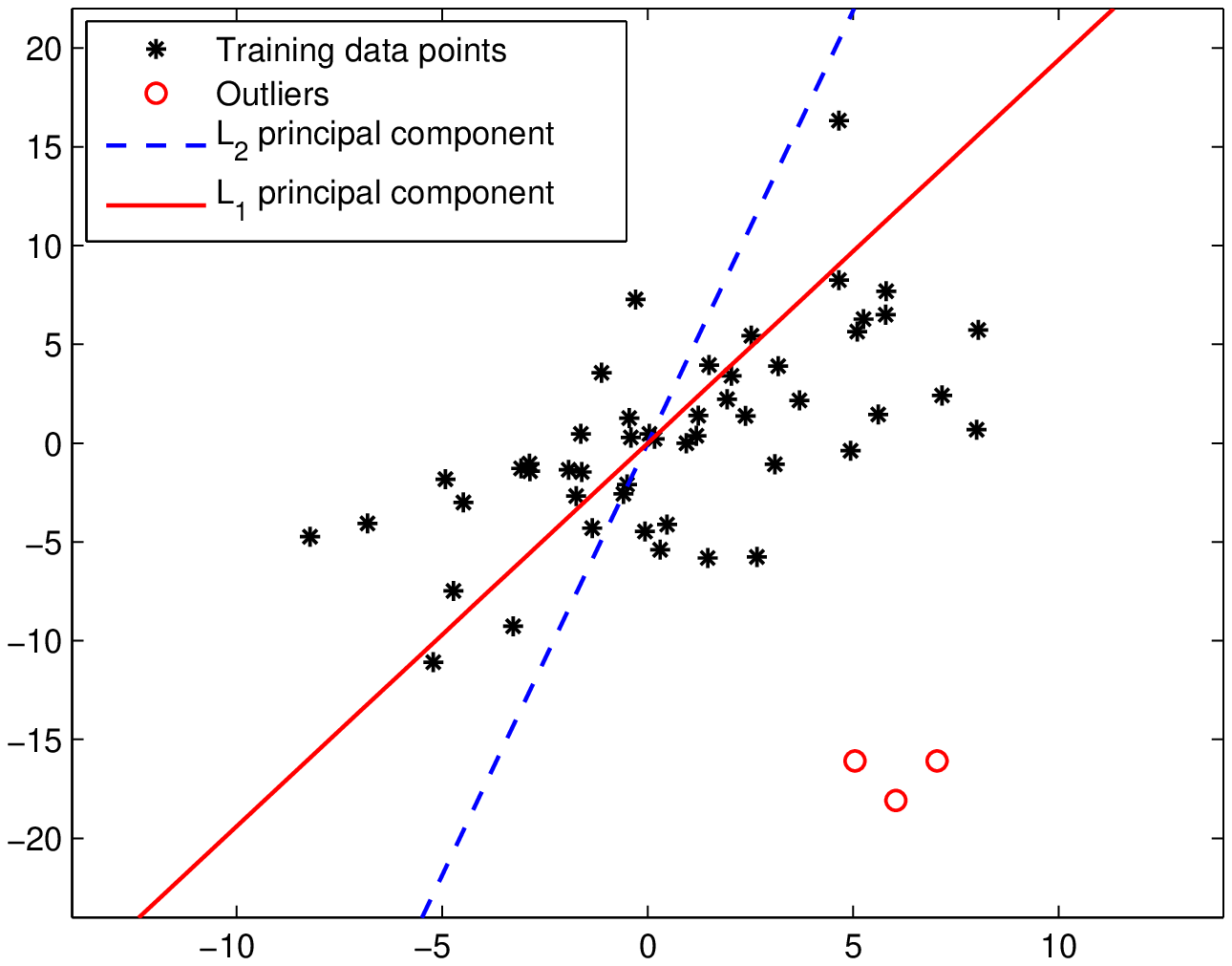}
\vspace{-0.8cm}
\caption{ }
\label{fig:dr2}
\end{subfigure}
\hfill
\begin{subfigure}[]{0.329\linewidth}
\centering
\includegraphics[width=\linewidth]{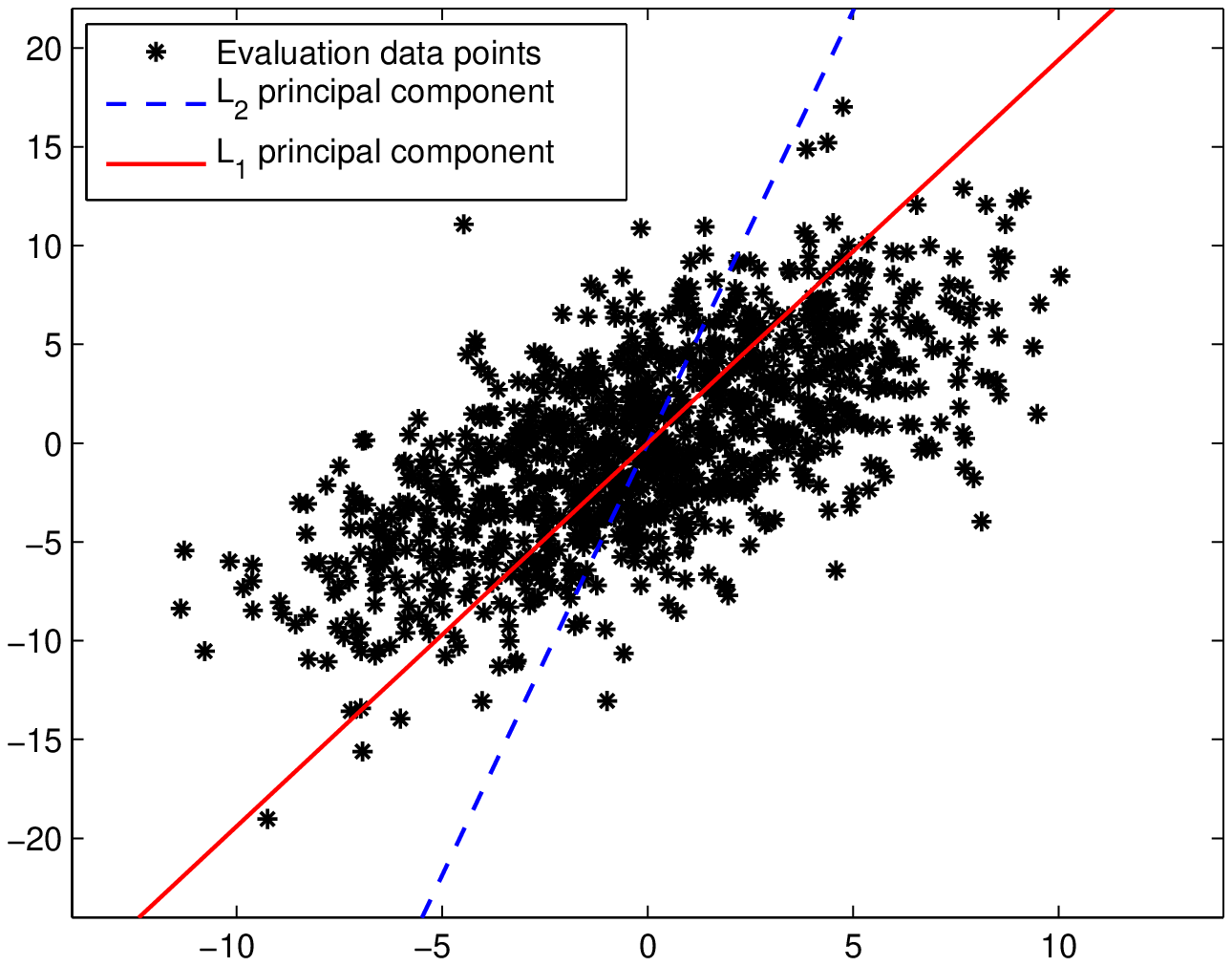}
\vspace{-0.8cm}
\caption{}
\label{fig:dr3}
\end{subfigure}
\vspace{-0.2cm}
\caption{
(a) Training data matrix $\mathbf X_{2 \times 50}$ with its $L_1$ and $L_2$ principal components ($K=1$).
(b) Training data matrix $\mathbf X_{2 \times 50}$ corrupted by three additional outlier points in bottom right with recalculated $L_1$ and $L_2$ principal components.
(c) Evaluation data set of $1000$ nominal points against the outlier infected (Fig. \ref{fig:dr2}) $L_1$ and $L_2$ principal components.}
\vspace{-0.5cm}
\label{fig:DR}
\end{figure*}

\subsection*{Experiment 1 - Data Dimensionality Reduction}

We generate a data-set $\mathbf X_{D \times N}$ of $N=50$  two-dimensional ($D=2$)  observation points  drawn from the Gaussian distribution $\mathcal N \left( \mathbf 0_2, 
\begin{bmatrix}
15 & 13 \\ 13 & 26
\end{bmatrix}
\right)$  as seen in Fig. \ref{fig:dr1}.
We calculate the $L_2$ (by standard SVD) and $L_1$ (by Section III.C, Case $d=2$, complexity about $50 \log_2 50$) principal component of the data matrix $\mathbf X$.%
\footnote{We note that without the presented algorithm, computation of the $L_1$ principal component of $\mathbf X_{2 \times 50}$ would have required complexity proportional to $2^{50}$ (by \eqref{eq:rank2}), which is of course infeasible.}
Then, we assume that our data matrix is corrupted by three outlier measurements, $\mathbf o_1, \mathbf o_2, \mathbf o_3$, shown in the bottom right corner of Fig. \ref{fig:dr2}.
We recalculate the $L_2$ and $L_1$ principal component of the corrupted data matrix $\mathbf X^{\text{CRPT}} = [\mathbf X, \mathbf o_1, \mathbf o_2 , \mathbf o_3]$ and notice (Fig. \ref{fig:dr1} versus Fig. \ref{fig:dr2}) how strongly the $L_2$ component responds to the outliers compared to $L_1$.
To quantify the impact of the outliers, in Fig. \ref{fig:dr3} we generate $1000$ new independent evaluation data points from $\mathcal N \left( \mathbf 0_2, 
\begin{bmatrix}
15 & 13 \\ 13 & 26
\end{bmatrix}
\right)$
and estimate the mean square-fit-error  $\text{E} \left\{ \|\mathbf x - \mathbf r\mathbf r^T \mathbf x \|_2^2 \right\}$ when $\mathbf r=\mathbf r_{L_2}(\mathbf X^{\text{CRPT}})$ or $\mathbf r_{L_1}(\mathbf X^{\text{CRPT}})$.
We find $\text{E}\left\{ \| \mathbf x  - \mathbf r_{L_2}(\mathbf X^{\text{CRPT}})\mathbf r_{L_2}(\mathbf X^{\text{CRPT}})^T \mathbf x \|_2^2 \right\} = 10.1296$ versus $\text{E}\left\{ \| \mathbf x  - \mathbf r_{L_1}(\mathbf X^{\text{CRPT}})\mathbf r_{L_1}(\mathbf X^{\text{CRPT}})^T \mathbf x \|_2^2\right\} =6.8387$.
In contrast, when the principal component is calculated from the clean training set, $\mathbf r= \mathbf r_{L_2}(\mathbf X)$ or $ \mathbf r_{L_1}(\mathbf X)$, we find mean square-fit-error $6.3736$ and $6.4234$, correspondingly.
We conclude that dimensionality reduction by $L_1$ principal components may loose only  little  in mean-square fit compared to $L_2$ when the designs are from clean training sets, but can protect significantly from outlier corrupted training.

\subsection*{Experiment 2 - Direction-of-Arrival Estimation} 

\begin{figure}[t!]
\centering
\includegraphics[width=1\linewidth]{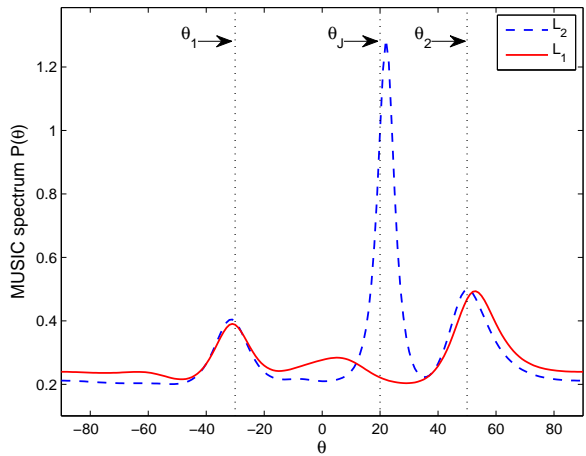}
\vspace{-0.8cm}
\caption{MUSIC power spectrum with $K=2$ $L_2$ or $L_1$ calculated principal components (data set of $N=10$ measurements with signals at $\theta_1=-30^\circ$ and $\theta_2 = 50^\circ$ of which one measurement is additive-jammer corrupted with $\theta_J=20^\circ$; $\text{SNR}_1 = 2$dB; $\text{SNR}_2 = \text{SNR}_J = 3$dB).}
\vspace{-0.3cm}
\label{fig:DOA}
\end{figure}

We consider a uniform linear antenna array of $D=5$ elements that takes $N=10$ snapshots of two incoming signals with angles of arrival  $\theta_1 = -30^\circ$ and $\theta_2 = 50^\circ$, 
\begin{align}
\mathbf x_n= A_1  \mathbf s_{\theta_1} + A_2   \mathbf s_{\theta_2} + \mathbf n_n, ~n=1, \cdots, 10,  \label{cleanmes}
\end{align}
where $A_1, A_2$ are the received-signal amplitudes with array response vectors $\mathbf s_{\theta_1}$ and $\mathbf s_{\theta_2}$, correspondingly, and 
$\mathbf n \sim \mathcal {CN}\left(\mathbf 0_{5}, \sigma^2 \mathbf I_{5} \right)$ is  additive white complex Gaussian noise.
We assume that the signal-to-noise ratio (SNR) of the two signals is  $\text{SNR}_1=10 \log_{10}\frac{A_1^2}{\sigma^2}\text{dB}=2\text{dB}$ and $\text{SNR}_2=10 \log_{10}\frac{A_2^2}{\sigma^2}\text{dB}=3\text{dB}$.
 Next, we assume that one arbitrarily selected measurement  out of the ten  observations $\mathbf X_{5 \times 10}=[\mathbf x_1, \cdots, \mathbf x_{10}] \in \mathbb C^{5 \times 10}$ is corrupted by a jammer operating at angle  $\theta_J= 20^\circ$ with amplitude $A_J = A_2$.
We call the resulting corrupted observation set  $\mathbf X^{\text{CRPT}} \in \mathbb C^{5 \times 10}$ and create the real-valued version  $\tilde{\mathbf X}^{\text{CRPT}} = [ \text{Re} \{\mathbf X^{\text{CRPT}}\}^T, ~ \text{Im} \{\mathbf X^{\text{CRPT}}\}^T ]^T \in \mathbb R^{10 \times 10}$ by $\text{Re}\{ \cdot\}, \text{Im}\{ \cdot\} $ part concatenation.
We calculate the $K=2$ $L_2$-principal components of $\tilde{\mathbf X}^{\text{CRPT}}$, $\mathbf R_{L_2} = [\mathbf r_{L_2}^{(1)}, \mathbf r_{L_2}^{(2)}] \in \mathbb R^{10 \times 2}$, and the $K=2$ $L_1$-principal components of $\tilde{\mathbf X}^{\text{CRPT}}$, $\mathbf R_{L_1} = [\mathbf r_{L_1}^{(1)}, \mathbf r_{L_1}^{(2)}] \in \mathbb R^{10 \times 2}$.
In Fig. \ref{fig:DOA}, we plot the standard $L_2$ MUSIC spectrum \cite{schmidt}
\begin{align}
P(\theta) \overset{\triangle}{=} \frac{1}{ \tilde{\mathbf s}_{\theta}^T (\mathbf I_{2D}- \mathbf R_{L_2}  \mathbf R_{L_2}^T) \tilde{\mathbf s}_{\theta}}, ~\theta \in \left( -\frac{\pi}{2}, \frac{\pi}{2}\right),
\end{align}
where  $ \tilde{\mathbf s}_{\theta} = [ \text{Re}\{{\mathbf s}_{\theta}\}^T, ~ \text{Im}\{{\mathbf s}_{\theta}\}^T ]^T$, as well as what we may call ``$L_1$ MUSIC spectrum'' with $\mathbf R_{L_1}$ in  place of $\mathbf R_{L_2}$.
 It is interesting to observe how $L_1$ MUSIC (in contrast to $L_2$ MUSIC) does not respond to the one-out-of-ten outlying jammer value in the data set and shows only the directions of the two actual nominal signals.

\section{Conclusions}

We presented for the first time in the literature optimal (exact) algorithms for the calculation of the maximum-$L_1$-projection component of data sets with complexity polynomial in the sample support size (and exponent equal to the data dimension).
We generalized to multiple $L_1$-max-projection components and presented an explicit optimal $L_1$ subspace calculation algorithm in the form of matrix nuclear-norm evaluations.
When $L_1$ subspaces are calculated on nominal ``clean'' training data, they differ little --arguably-- from their $L_2$-subspace counterparts in least-squares fit.
However, subspaces for data sets with possibly erroneous,   ``outlier'' entries,   $L_1$ subspace calculation offers significant robustness/resistance to the presence of inappropriate data values.

\end{document}